\documentclass[letter,11pt]{article}
\usepackage[letterpaper, total={6.5in, 9in}]{geometry}
\usepackage{hyperref}\usepackage{breakurl}
\usepackage[
    backend=biber,
    sortlocale=en,
    natbib=true,
    url=false, 
    doi=true,
    eprint=true,
]{biblatex}
\renewcommand{\citet}[1]{\citeauthor{#1}\,\autocite{#1}}
\let\cite=\autocite 
\usepackage[english]{babel}
\usepackage{graphicx, csquotes}\MakeOuterQuote{"}
\usepackage{versions}\excludeversion{ignore}
\usepackage[leftmargin=1em,rightmargin=4em]{quoting}
\author{Rakesh M. Verma 
  \\ Department of Computer Science \\ University of Houston \\ \texttt{rmverma2@central.uh.edu} \\
  \and
  Nachum Dershowitz 
\\ School of Computer Science \\ Tel Aviv University \\ \texttt{nachum@tau.ac.il} \\
  }
  
  \title{\bf The Pitfalls of Publishing in the Age of LLMs:\\ Strange and Surprising Adventures with a High-Impact NLP Journal}
\begin{document}

\maketitle
\begin{abstract}
We show the fraught side of the academic publishing realm and illustrate it through a recent case study with an NLP journal. 
\end{abstract}

\section{Introduction}
In the dawn of the age of Large Language Models (LLMs), already much has been said about how researchers are making use of LLMs to author articles. 
For example, according to an article in \textit{Scientific American} \cite{sciam}, ``One percent of scientific articles published in 2023 showed signs of generative AI’s potential involvement, according to a recent analysis.''
However, far less has been said about how reviewers are now abusing their role, sometimes with the editor's collusion. 

Here is our report of a case in point. We submitted a manuscript on domain-independent deception detection to a highly respected journal. 
As a consequence of a reviewer's use of an LLM, we both received a most peculiar review and also lost the promised confidentiality regarding our submission. 

\section{The Strange Case of a  Submission in Computational Linguistics}

On the last day of August 2023, we, along with three other colleagues, submitted an article to a prominent journal in natural language processing (NLP) slash computational linguistics (CL).
Two and a half months later, on 17 November,
the editor-in-chief sent us three reviews.
One of them stood out, as it was clearly written  for the most part by a machine.
We quote it verbatim below:
\begin{quotation}
\noindent
Reviewer B:
\begin{enumerate}\setlength{\itemindent}{-10pt}
\item[1.] Originality:
Based on the information provided in the paper, the paper does contain some
new information and insights related to the field of deception detection.
However, the paper presents some novel methods and findings, it's essential
to assess whether these contributions significantly advance the current
state of the field.
\item[2.] Relationship to Literature:
\begin{itemize}
\item[--]The paper demonstrates a reasonable understanding of the relevant
literature in the field. It cites a wide range of literature sources,
including academic papers, books, and software tools, to provide context and
support for the research. The references cover various aspects of deception
detection, including linguistic analysis, machine learning, and psychology,
which indicates a comprehensive review of the literature.
\item[--]The paper generally acknowledges and cites significant work in the field
of deception detection. It references various established techniques,
datasets, and prior research contributions related to the detection of
deceptive content. However, it's essential to note that the significance of
the work can vary depending on the specific focus of the paper. While the
paper references a substantial body of literature, it may not cover every
single significant work due to the breadth of the field.
Overall, the paper's literature review appears to be thorough, covering a
wide range of relevant sources. However, the assessment of whether any
"significant" work is ignored can be subjective, as it depends on the
specific criteria for significance and the scope of the paper. We may have
varying opinions on whether certain works should have been included, but the
paper generally provides a solid foundation in terms of referencing relevant
literature.
\end{itemize}

\item    [3.] Methodology
\item[--]The paper does build its argument on an appropriate base of theory and
concepts related to deception detection. It draws upon established theories
and techniques from linguistics, natural language processing, and psychology
to frame its research. However, while the theoretical foundation is
generally sound, there could be more explicit connections between the chosen
theories and the specific research questions addressed in the paper.
\item[--]The design of the research in the paper is generally well-structured. It
presents a systematic exploration of various methods and techniques for
deception detection, including linguistic analysis, machine learning, and
cross-domain applications. However, the paper lacks some details on specific
aspects of the research design, such as the selection criteria for datasets
and the rationale behind the chosen evaluation metrics. Providing more
transparency in these areas would enhance the paper's rigor.
 \item[--]The paper employs a wide range of methods, including linguistic analysis,
machine learning, and cross-domain experiments. These methods are suitable
for investigating deception detection, given the multidisciplinary nature of
the field. However, the appropriateness of the methods could be further
validated by discussing their advantages and limitations in the context of
the research questions. Additionally, the paper should clarify the criteria
used to select specific methods or models for experimentation and analysis.
In summary, while the paper does build its argument on relevant theories and
concepts and demonstrates a well-structured research design, there is room
for improvement in explicitly connecting the chosen theories to the research
questions and providing more details on the research design and method
selection criteria.

\item    [4.] Results:
The presentation of results and the analysis in the paper generally follow a
structured and systematic approach. However, there are areas for improvement
in terms of clarity and the alignment of conclusions with the other elements
of the paper. Here's a detailed evaluation:
\item[--]Strengths:

a. The paper organizes the presentation of results in a systematic manner,
often following a clear structure of introducing the research question,
detailing the methodology, and presenting findings.

b. Visual aids, such as tables, figures, and charts, are used to illustrate
key results, which enhances the clarity of the presentation.

c. The paper incorporates quantitative analysis, which is appropriate for
the research objectives, and uses statistical measures to describe and
evaluate results.

d. In some sections, the paper compares its findings with existing methods
or benchmarks, which provides context and helps assess the significance of
the results.

\item[--]Areas for Improvement:

a. While the paper presents results, the depth of analysis could be
enhanced. In some instances, the analysis is somewhat descriptive and lacks
deeper insights or interpretations of the findings.

b. The paper should clearly indicate the statistical significance of the
results where applicable. This is essential to support the validity of the
findings.

c. For a more comprehensive understanding, qualitative analysis (e.g.,
linguistic analysis of deceptive cues) could be incorporated where relevant,
especially in sections related to linguistic aspects of deception.

\item[--]Conclusions (Strengths and Areas for Improvement):

a. The conclusions section provides a concise summary of the key findings
and their implications, which is a strength.

b. The conclusions should be scrutinized to ensure they align closely with
the research objectives and the results presented earlier in the paper. In
some cases, there may be a need for clearer connections between results and
conclusions.

c. It would be beneficial for the paper to include recommendations for
future research based on the findings. This helps to demonstrate the broader
significance of the research.

d. While the paper discusses theoretical aspects, it could provide more
insights into the practical implications of the research. How can the
findings be applied in real-world scenarios or specific domains?

e. The conclusions should acknowledge the limitations of the research. This
adds transparency and helps readers assess the generalizability of the
findings.

In summary, the paper generally presents results in a structured manner, but
there is room for deeper analysis and more explicit connections between
results and conclusions. Strengthening the alignment with research
objectives, discussing practical implications, and addressing limitations
will enhance the overall quality of the conclusions section.

\item    [5.] Implications for research, practice and/or society:
\item[--]The paper extensively reviews and discusses existing literature and
techniques related to deception detection, providing a comprehensive
overview of the field. This can serve as a valuable resource for researchers
and practitioners looking to understand the state of the art.
\item[--]It introduces various methodologies and approaches for deception
detection, including linguistic analysis, machine learning, and multi-modal
techniques. These methods have the potential to impact research in natural
language processing, machine learning, and psychology.
\item[--]The discussion of datasets and evaluation metrics can guide researchers in
selecting appropriate resources for their studies and help standardize
evaluation practices in the field, contributing to the improvement of
deception detection systems.
\item[--]While the paper doesn't explicitly state practical applications, the
techniques discussed, if further developed, could have real-world
implications. For instance, improving deception detection algorithms could
enhance security measures in various domains, including cybersecurity and
law enforcement.
\item[--]Overall, the paper indirectly identifies implications for research and
practice by presenting a wealth of information and methodologies related to
deception detection. However, it could benefit from more explicit
discussions of the practical applications and societal impacts of these
techniques.
 \item[--]To further bridge the gap, the paper could benefit from case studies or
practical examples illustrating how these techniques have been applied in
real-world scenarios, demonstrating their practical utility.

In summary, the paper presents a comprehensive overview of deception
detection techniques and research, indirectly identifying implications for
research and practice. While it establishes a strong theoretical foundation,
it could enhance its impact by explicitly discussing practical applications
and providing concrete examples of how these techniques can be used in
real-world contexts.

\item    [6.] Quality of Communication:
\begin{itemize}
    
\item[--]The paper generally maintains a high level of clarity in its expression.
It effectively conveys complex technical concepts, making them accessible to
readers with the expected knowledge in the field.
\item[--]Sentence structures are generally clear and coherent, aiding in the
comprehension of the content. However, there are instances of overly long
sentences and complex language that may pose readability challenges for some
readers. Streamlining sentence structure and using simpler language in these
cases would enhance overall readability.
\item[--]The paper shows a reasonable level of attention to clarity and
readability, evident through well-structured sections, headings, and
subheadings that guide readers through the content.
\item[--]Acronyms and technical terms are typically defined upon first use,
enhancing comprehension for a broad readership.
\item[--]While there are some grammatical issues and typographical errors present
in the text, they do not significantly detract from the overall clarity. For
examples:

a. Long sentences:
\begin{trivlist}\itemsep 5pt
\item Original   Sentence:       "In     this    section,        we      provide a comprehensive overview of
the existing literature on deception detection, highlighting the key
approaches, methodologies, and challenges in this domain, and we discuss how
the proposed task aligns with and contributes to the broader research
landscape."

Revised Sentence: "In this section, we comprehensively review existing
literature on deception detection. We highlight key approaches,
methodologies, and challenges in this domain. Additionally, we discuss how
the proposed task aligns with and contributes to the broader research
landscape."

\item
    
Original   Sentence: "One of the central challenges in deception detection is
the development of robust models that can effectively generalize across
different domains, languages, and modalities, as deceptive content can
manifest in various forms, including textual, visual, and audiovisual
media."

Revised Sentence: "A central challenge in deception detection is developing
robust models that generalize effectively across diverse domains, languages,
and modalities. Deceptive content can take various forms, such as textual,
visual, and audiovisual media."

\item  Original   Sentence: "Furthermore, the task of deception detection often
involves addressing issues related to imbalance in the distribution of
deceptive and truthful instances, as deceptive content is typically less
prevalent than truthful content, which can lead to biased models that favor
the majority class."

Revised Sentence: "Moreover, deception detection tasks frequently require
addressing issues related to imbalanced distributions of deceptive and
truthful instances. Deceptive content is usually less common than truthful
content, potentially leading to biased models favoring the majority class."

\item  Original   Sentence: "To address these challenges, researchers have explored a
wide range of features and representations, including linguistic features,
stylistic cues, syntactic patterns, semantic embeddings, visual attributes,
acoustic features, and multimodal fusion techniques, to capture the nuances
of deceptive communication."

Revised Sentence: "Researchers have tackled these challenges by exploring
diverse features and representations. These include linguistic features,
stylistic cues, syntactic patterns, semantic embeddings, visual attributes,
acoustic features, and multimodal fusion techniques, all aimed at capturing
the nuances of deceptive communication."

\item  Original   Sentence: "Additionally, recent advances in deep learning and
neural network architectures have shown promise in improving the accuracy of
deception detection models, particularly in tasks that involve complex data,
such as image and speech analysis, and the integration of contextual
information from social media platforms."

Revised Sentence: "Furthermore, recent advancements in deep learning and
neural network architectures show promise in enhancing the accuracy of
deception detection models. This is especially notable in tasks involving
complex data, like image and speech analysis, as well as in integrating
contextual information from social media platforms."

\item  Original   Sentence: "The task of authorship attribution, which aims to
determine the most likely author of a given text based on linguistic
patterns, has been extensively studied in the field of natural language
processing and computational linguistics, and it shares similarities with
deception detection in terms of the linguistic analysis of text."

Revised Sentence: "Authorship attribution, a task that seeks to identify the
most probable author of a text through linguistic patterns, has garnered
extensive research attention within the realms of natural language
processing and computational linguistics. This task bears similarities to
deception detection due to its reliance on linguistic text analysis."

\item  Original   Sentence: "While the primary objective of authorship attribution is
to identify the authorship of a text, it also has applications in deception
detection, as deceptive content may exhibit unique linguistic traits that
can be traced back to specific authors."

Revised Sentence: "Although the primary goal of authorship attribution is
text authorship identification, it finds applications in deception detection
as well, given that deceptive content often displays distinct linguistic
characteristics attributable to particular authors."

\item  Original   Sentence: "In this section, we delve into the various techniques
and methods employed in authorship attribution, exploring both traditional
and state-of-the-art approaches to provide a comprehensive understanding of
the field."

Revised Sentence: "In this section, we explore a wide array of techniques
and methods used in authorship attribution, encompassing both conventional
and cutting-edge approaches to offer a thorough grasp of the field."

\item  Original   Sentence: "Furthermore, we investigate the challenges and
limitations associated with authorship attribution, highlighting issues
related to cross-domain and cross-lingual generalization, data availability,
and privacy concerns."

Revised Sentence: "Additionally, we examine the challenges and constraints
linked to authorship attribution. We emphasize difficulties related to
cross-domain and cross-lingual generalization, data availability, and
privacy considerations."

\item  Original   Sentence: "Finally, we draw connections between authorship
attribution and deception detection, illustrating how insights from the
former can inform the latter, especially in tasks involving linguistic
analysis and text-based communication."

 "Lastly, we establish connections between authorship
attribution and deception detection, demonstrating how insights from the
former can provide valuable input to the latter, particularly in tasks that
entail linguistic analysis and text-based communication."

\item *These revisions aim to simplify complex sentences by breaking them into
smaller, more focused segments, enhancing readability and comprehension

\end{trivlist}

\item[--]Here are examples of some grammatical errors:
\begin{itemize}
    
\item[$>$] "Deception detection is a significant area of research that has gained
increasing attention in recent years," consider rephrasing to "Deception
detection is a significant area of research that has garnered increasing
attention in recent years.". In the same sentence, consider adding "the"
before "recent years" to make it "in the recent years."
\item[$>$] "This paper presents an overview of deception detection, focusing on
common techniques that have been widely utilized and the challenges
associated with this field," consider rephrasing for clarity: "This paper
presents an overview of deception detection, with a focus on common
techniques that have been widely utilized and the challenges associated with
this field."
\item[$>$] "The aim of this paper is to provide insights into the key approaches and
trends in deception detection," you might consider adding "the" before
"deception detection" to make it "in deception detection."
\item[$>$] "The past decade has seen significant advancement in deception detection
research," consider changing "advancement" to "advancements" for plural
agreement.
\item[$>$] "Related research is reviewed to provide a comprehensive understanding of
the field," consider specifying what research is being referred to for
clarity. For example, "Existing research is reviewed to provide a
comprehensive understanding of the field."

\item[$>$] "In the next sections, the paper discusses the major techniques and
approaches in deception detection," you can make it more concise by saying,
"In the following sections..."
\item[$>$] "Additionally, the paper highlights the importance of deception detection
in various domains," you can improve clarity by specifying which domains are
being referred to. For example, "Additionally, the paper highlights the
importance of deception detection in various domains such as..."
\item[$>$] "These models have been trained on large datasets consisting of both
deceptive and truthful content," consider adding "which" before "consisting"
for smoother sentence flow: "These models have been trained on large
datasets, which consist of both deceptive and truthful content."
\item[$>$] "This allows them to learn patterns and features indicative of deception," you can improve clarity by specifying what "them" refers to: "This allows these models to learn patterns and features indicative of deception."
\item[$>$] 
"Additionally, it is crucial to ensure that these models are not only
accurate but also fair and unbiased," consider specifying what "these
models" refer to for clarity: "Additionally, it is crucial to ensure that
machine learning models used in deception detection are not only accurate
but also fair and unbiased."
\item[$>$] "Feature engineering involves selecting and transforming relevant features
to improve model performance," consider specifying what "Feature
engineering" is referring to: "Feature engineering involves selecting and
transforming relevant features within the dataset to improve model
performance."
\item[$>$] "In this section, we will overview some of the key challenges associated
with deception detection," the word "overview" should be replaced with
"provide an overview of" for grammatical correctness: "In this section, we
will provide an overview of some of the key challenges associated with
deception detection."
\item[$>$] "Deceptive behavior is not confined to written or spoken language,"
consider specifying what "Deceptive behavior" refers to: "Deceptive behavior
in various contexts is not confined to written or spoken language."
\item[$>$] "One of the central challenges is the lack of a universal definition of
deception," consider specifying what "One" refers to: "One of the central
challenges in deception detection is the lack of a universal definition of
deception."
\item[$>$] "Another challenge is the dynamic nature of deception," consider
specifying what "Another" refers to: "Another challenge in deception
detection is the dynamic nature of deception."
\item[$>$] "Research on deception detection spans various domains, including
linguistics, psychology, and computer science," consider using "spans" for
consistency: "Research on deception detection spans various domains,
including linguistics, psychology, and computer science."
\item[*]Please note that these are relatively minor grammatical suggestions, and
the text is generally well-structured.
\end{itemize}
\end{itemize}
\end{enumerate}

In summary, the paper generally succeeds in expressing its case with
technical language relevant to the field and the expected knowledge of the
journal's readership. However, there is room for improvement in terms of
sentence structure and language simplification, which would enhance overall
readability without compromising technical rigor. Additionally, careful
proofreading to eliminate grammatical errors and typos would further enhance
clarity.
\end{quotation}

As it was obvious to us that the above review was written mainly by a chatbot, we wrote on 1\nolinebreak\@ December to the editor-in-chief, as follows:%
\footnote{Names have been replaced to maintain anonymity of the parties concerned.}

\medskip
\fbox{\begin{minipage}
{\linewidth}
\begin{quoting}
\noindent
Dear Prof. N.,
 
\smallskip
Thank you very much for considering our manuscript for [your journal].
 
\smallskip
We were, however, horrified to observe that B’s review is blatantly produced mostly by a Chatbot.  
Consequently, many of the comments are superficial, misguided, or unwarranted and couched in apologia.
 
\smallskip
Ironic, considering that the submission was about deception. In any case, to our mind, this should be totally unacceptable. We trust you to thoroughly investigate this incident and take appropriate action against the party or parties responsible for disregarding the fundamental principles of scientific review.
 
\smallskip
Please do not hesitate to contact us if you need any help with the investigation and let us know if you have any questions.

\smallskip
Kind Regards,
 
\smallskip
Rakesh Verma (On behalf of all the authors)

\end{quoting}
\end{minipage}
}

\bigskip
After sending a reminder on 16 January 2023, we received the following response from one of the editors:

\medskip
\fbox{\begin{minipage}
{\linewidth}
\begin{quoting}
\noindent
Dear Rakesh,
 
\smallskip
Thank you for bringing this to our attention. And apologies for the delayed response.
 
\smallskip
Using an automated language generation tool for writing reviews is unacceptable. Such conduct will not be tolerated and they have been made aware of this. Further, they have been told that they will not be invited to review \dots\@ papers in the future.
 
\smallskip
Reviewer B is not a standing reviewer for [this] journal. I had chosen them because they had written one of the papers cited by your submission and had done related work. But as a result of this review, they will be excluded from any future reviewing role in [this] journal.
 
\ldots
 
\smallskip
Best wishes.\\
S.
\end{quoting}
\end{minipage}
}

\bigskip
We wrote back saying

\medskip
\fbox{\begin{minipage}
{\linewidth}
\begin{quoting}
\noindent
Dear S.,
 
Thank you very much for your response, but we were quite disappointed to learn that:

\begin{itemize}
\item[a)] the editor in charge of our paper did not even notice this issue and just forwarded the fake review,
\item[b)] there does not seem to be any plan to prevent a recurrence of such a problem, and
\item[c)] there is no mechanism in place to stop said reviewer (or others like them) from repeating this offense with another journal or conference in the future
\end{itemize}

Sincerely,\\
Rakesh Verma
\end{quoting}
\end{minipage}
}

\bigskip
In response, the editor-in-chief of the journal, wrote:

\medskip
\fbox{\begin{minipage}
{\linewidth}
\begin{quoting}
\noindent
Dear Rakesh,
\smallskip

FYI: The \dots\@ Executive Committee is urgently setting up a publication ethics committee to develop a detailed policy about misconduct in peer review (e.g., using LLMs to generate and submit text that looks like a peer review), including detection, imposition of penalty, etc. 
[The policy] will encompass \dots\@ both journal and conference publications.

\smallskip\noindent
Regards\\
N.
\end{quoting}
\end{minipage}
}

\bigskip
Finally:

\fbox{\begin{minipage}
{\linewidth}
\begin{quoting}
\noindent
Dear N.,
\smallskip

Thank you very much for the information. 

However, it is cold comfort for us, since we have lost all the confidentiality expectations that we had of our paper. As OpenAI clearly states, any information uploaded to ChatGPT will be used as training data. 

\dots

Best regards,
Rakesh
\end{quoting}
\end{minipage}
}

\section{Experiments with ChatGPT4}
We have received no further response from the editor. Based on  one usable review (the third review was specious), a decision to reject our manuscript was taken with lightening speed in the world of academia.
(Having  lost all confidentiality expectations, we decided to upload our paper to Arxiv~\cite{vermaD2024}.)

As is obvious from the review, it could not possibly have been authored by a human reviewer. For example, the review consists of inane suggestions in the form of offering our sentence from the manuscript as an improved form of our sentence from the manuscript.

So we then uploaded our manuscript to ChatGPT4 from a premium account and also the review in question and we asked ChatGPT4 to act as an expert reviewer with a decade of experience and to rate various aspects of the review on a scale of 1 to 5. We did get back some useful comments and scores. 
However, no amount of hints, such as what is wrong with this review or whether there are ethical considerations that you have missed, etc., led ChatGPT4 to assert that this review might have been generated unethically by an AI chatbot. 
Rather, it concluded:
\begin{quotation}
Based on the review content provided, there are no evident ethical concerns regarding confidentiality breaches, lack of constructiveness, bias, or unprofessionalism. The critique seems to have been conducted in a manner that aligns with the ethical standards expected in scholarly peer review, focusing on improving the paper within the confines of the journal’s confidentiality policy.

In summary, the review, as described, adheres to the ethical and professional standards expected in the academic peer review process. It provides constructive, objective, and professional feedback without breaching the journal's confidentiality policy or compromising the integrity and fairness of the review.    
\end{quotation}

\section{Conclusions}

So, we have reached the age of maximum peril for authors and reviewers alike. How have we gotten here? By just trying to predict the next word given preceding words. As they say, all technology has unintended uses and unintended victims (along the lines of Robert K. Merton's 1936 ``Law of Unanticipated Consequences''). Clearly, LLMs lack self-awareness at the present time.

\printbibliography
\end{document}